\documentclass{article}

\usepackage[main, final]{neurips_2026_vericode}

\usepackage[utf8]{inputenc} 
\usepackage[T1]{fontenc}    
\usepackage{hyperref}       
\usepackage{url}            
\usepackage{booktabs}       
\usepackage{amsfonts}       
\usepackage{nicefrac}       
\usepackage{microtype}      
\usepackage{xcolor}         
\usepackage{booktabs}
\usepackage{amsmath}
\usepackage{tabularx}
\title{When Verification Is Not Agreement: Cross-Agent LLM-to-Dafny Formalization}

\author{%
  Priyanath Maji \\
  Georgia Institute of Technology\\
  Atlanta, GA 30332 \\
  \texttt{pmaji3@gatech.edu} \\
}

\begin{document}

\maketitle

\begin{abstract}
A verified Dafny postcondition certifies that a formula holds, but it does not guarantee that the result was reached through genuine independent reasoning. We introduce a cross-agent formalization protocol where a Specifier model translates clinical trial criteria into a Dafny predicate, and an Implementer model writes an operational method checked against it. A same-model control establishes that even with no capability gap, independent re-extraction from identical text agrees only 59.3\% of the time. Evaluating four model configurations on 96 clinical trial documents, we find that agreement is highly sensitive to how model capability is allocated across the two roles: the GPT-4.1-nano/GPT-4.1-mini configuration achieves 53.0\% agreement when the stronger model is the Implementer, but only 14.3\% when the roles are reversed. The strongest tested configuration, GPT-5.6-luna/GPT-5.4-mini, improves agreement to 34.4\% but remains below the same-model baseline and is dominated by direct contradiction (42.7\%). We also detect a critical verification artifact in which a weaker Implementer satisfies the postcondition by invoking the Specifier's predicate directly; after structural auditing, corrected agreement falls from 14.3\% to 13.2\%. Our results show that verification success alone is insufficient evidence of independent cross-agent agreement, and that model capability does not guarantee agreement when formalizing the same source independently.
\end{abstract}

\section{Introduction}
\label{sec:intro}

A verified Dafny~\cite{leino2010dafny} \texttt{ensures} clause establishes that the postcondition holds under the verifier's assumptions. It does not tell you \emph{why} it holds, or whether the two artifacts being checked against each other were produced through genuinely independent reasoning. A common pattern in neuro-symbolic pipelines pairs a Specifier agent, which produces a formal specification, with an Implementer agent, whose code is checked against that specification by a verifier. But verification success in this pattern is only meaningful evidence of agreement if the two agents' artifacts were actually derived independently. If both agents are shown the same pre-resolved facts before they begin, or if one agent can shortcut its own reasoning by referencing the other's output, verification succeeds trivially, and certifies nothing about independent agreement.

We study this distinction directly in a domain involving numeric eligibility criteria extracted from clinical trial text, chosen specifically because it admits a clean way to test it without requiring human-annotated gold labels. We first confirm, empirically, that the naive implementation of this pattern is vulnerable to exactly the failure it is not supposed to permit. When a Specifier and an Implementer are given an identical, already-parsed list of extracted facts, verification succeeds on the first attempt in 50 of 52 documents, and the mutation-adequacy guard~\cite{black2000mutation, cordy2023mutation} catches all synthetic mutants with zero escapes. This is not evidence the pipeline works; it is evidence the two agents were never in a position to disagree. We then redesign the protocol so that each agent reads the raw source text independently, sharing only the structural signature necessary for Dafny to type-check the Implementer's method against the Specifier's predicate, never the Specifier's actual thresholds. Genuine agreement must now be earned, and disagreement is diagnosable into distinct, interpretable failure modes, all without requiring human-annotated gold labels.

Two results emerge from applying this protocol across four Specifier/Implementer model configurations on 96 clinical trial documents from the Chia corpus~\cite{kury2020chia}. First, a same-model control shows that independent re-extraction from identical text agrees only 59.3\% of the time even with no capability gap, establishing that formalization agreement is a genuinely nontrivial task before any capability asymmetry is introduced. Second, agreement is highly sensitive to how model capability is allocated across the two roles. With GPT-4.1-nano and GPT-4.1-mini, agreement reaches 53.0\% when the stronger model serves as the Implementer, but only 14.3\% when the roles are reversed. The strongest tested configuration, GPT-5.6-luna/GPT-5.4-mini, achieves 34.4\% agreement but remains well below the same-model baseline and exhibits a qualitatively different failure profile dominated by direct contradiction. These results indicate that model capability alone does not guarantee agreement when independently formalizing the same source.

Finally, structural auditing exposes a concrete limitation of verification as an independence test. In the configuration with the weaker Implementer, one case classified as agreement was satisfied by invoking the Specifier's predicate directly rather than reconstructing the logic independently. After removing this shortcut, corrected agreement falls from 14.3\% to 13.2\%. This artifact would have passed the Dafny postcondition while violating the intended independence of the two-agent protocol, demonstrating that verification success alone cannot distinguish genuine reconstruction from a structurally dependent implementation.

Recent work has explored integrating formal verification into LLM-based code generation, including consistency checking between generated code and formal annotations, as well as verification-guided program synthesis and agent safety \cite{clover2024,leetproof2026,veriguard2025}. However, successful verification does not necessarily establish that independently generated formalizations agree on the meaning of the original natural-language requirements.

\paragraph{Contributions.}
We first provide a cross-agent evaluation protocol in which two models independently formalize the same source text, with their artifacts connected through a shared Dafny contract. This configuration allows genuine agreement, coverage gaps, and direct contradictions to be distinguished mechanically from the two agents' independently derived atom sets without requiring human-annotated gold labels. Second, we present empirical evidence via a same-model control that independent-formalization agreement is substantially below 100\% even absent any capability gap, establishing the baseline against which capability and role allocation must be interpreted. Third, we demonstrate that agreement is highly sensitive to how model capability is allocated across the Specifier and Implementer roles, where reversing roles produces a large change in agreement while the strongest tested configuration remains below the same-model baseline and exhibits a contradiction-dominated failure signature. Finally, we surface a concrete, structurally audited case of verification succeeding without genuine independent reconstruction, where an Implementer invokes the Specifier's predicate directly to satisfy Dafny's postcondition, demonstrating why verification success alone is insufficient to establish cross-agent agreement.

\section{Related Work}

\paragraph{LLM-based formalization and verification.}
Recent work has explored using large language models to translate natural-language intent into formal specifications and program assertions, including formal postconditions~\cite{endres2024nl2postcond} and specifications for verification-aware languages such as Dafny~\cite{lahiri2024evaluating}. Other work has studied LLM-assisted generation and repair of formally verified programs using verifier feedback~\cite{loughridge2025dafnybench}, as well as autoformalization into proof-assistant languages such as Lean~\cite{murphy2024leaneuclid}. These efforts primarily evaluate whether an individual generated artifact captures the intended behavior or can be made to satisfy a formal verifier. Our focus is different because we evaluate whether two agents independently deriving formal artifacts from the same natural-language source converge on the same formalization. Thus, verification is used not only to certify an artifact, but as part of an independence-sensitive agreement protocol.

\paragraph{Mutation testing and specification adequacy.}
Mutation analysis has long been used to assess the adequacy of software tests, and prior work has applied mutation operators directly to formal specifications~\cite{black2000mutation}. More recent work uses mutation-based analysis to assess and strengthen formal specifications by determining whether they distinguish deliberately introduced behavioral faults~\cite{phippatananunth2022mutations,cordy2023mutation}. We use the same general principle of mutation adequacy as a complementary audit of generated specifications. Our setting differs in that mutations are used to test the discriminative strength of LLM-generated specifications within a cross-agent formalization protocol, rather than primarily to evaluate a conventional test suite or manually authored specification.

\paragraph{Domain and clinical eligibility formalization.}
Clinical trial eligibility criteria have been studied extensively as a structured natural-language information extraction problem. The CHIA corpus provides annotated eligibility criteria representing semantic structures transformable into Boolean query logic~\cite{kury2020chia}, while frameworks like Criteria2Query 3.0 use large language models directly to convert criteria into structured representations for clinical trial matching~\cite{park2024criteria2query}. These systems demonstrate the feasibility of formalizing eligibility criteria at scale, but they primarily evaluate the resulting representations through downstream retrieval, matching, or execution performance. We instead treat eligibility criteria as a non-software specification where numerical conditions permit mechanically decomposing two independently generated formalizations into comparable logical atoms. This lets us evaluate whether the two derivations semantically agree with each other without requiring external gold labels. Unlike traditional verified refactoring or program equivalence tools, the artifacts being compared are independently derived from natural-language source text rather than produced through a known program transformation.

\section{Method}
\label{sec:method}

\subsection{Task and Schema}

We work with eligibility-criteria text from the Chia corpus~\cite{kury2020chia}. For each document, we concatenate the inclusion and exclusion criteria passages into a single source text. Both agents in a pairing are restricted to a shared vocabulary of variable names, together with a type annotation (integer vs.\ real) for each, defined in a fixed schema file. Restricting both agents to the same approved names is what makes independently-produced atom sets comparable after the fact. Neither agent is free to invent an idiosyncratic name for a concept the schema already covers, so if two agents both correctly detect a criterion, they are expected to instantiate it under the same variable name.

\subsection{Independent Dual-Route Extraction}

Each document is extracted twice, independently, directly from the raw source text, never from a pre-parsed or pre-resolved fact list. The Specifier uses a \emph{literal} extraction prompt (Route A), instructed to preserve the exact boundary and strictness of the source wording. The Implementer uses an \emph{interpretive} extraction prompt (Route B), instructed to reason about clinical intent and infer thresholds that are implied but not explicitly numeric in the text. Both prompts share an identical format contract where every emitted value must reduce to a single explicit comparison operator and numeric threshold, or a two-sided \texttt{A and B} range. Purely categorical criteria with no numeric bound are skipped, and only vocabulary-approved variable names may be used. The two roles are served by separate API clients with no shared context, ensuring the Implementer's extraction cannot be conditioned on anything the Specifier produced.

Raw model output is parsed into a list of \texttt{SimpleAtom(var, op, value)} triples via a single canonical normalizer. Two-sided ranges are split into two atoms sharing a variable. Any value that fails to match the canonical inequality grammar, or names a variable outside the approved vocabulary, is routed to an out-of-vocabulary report rather than silently dropped, ensuring that extraction failures remain diagnosable rather than invisible.

\subsection{Cross-Agent Formalization}

The Specifier's atoms are conjoined into a Dafny predicate \texttt{MatchSpec}, taking one parameter per distinct variable appearing in its own atom set. The Implementer's prompt contains exactly two document-specific inputs, specifically the required parameter signature containing names and Dafny types only derived from the Specifier's atom set, alongside the Implementer's own independently extracted atoms from Route B, which it is instructed to use as the basis for its logic. The Specifier's actual thresholds, operators, and predicate body are never included in this prompt. The Implementer then writes an operational Dafny method \texttt{CheckEligibility} whose body is derived from its own atoms, with a postcondition \texttt{ensures eligible == MatchSpec(\ldots)} checked by the Dafny verifier.

Absent a verification shortcut, successful verification requires the Implementer's independently derived logic to satisfy the Specifier's predicate. The protocol therefore treats first-attempt verification as apparent agreement, which is subsequently subjected to the mutation and verification-gaming audits described below. For the runs reported here, verification is evaluated once per document with no repair retry. The implementation supports a bounded repair loop in which a Symbolic Repair Agent revises only the Implementer's method against the Dafny counterexample, but this was not exercised in the results below.

\subsection{Outcome Categorization}

Verification success or failure alone conflates several qualitatively different outcomes, so we decompose each document's result into a category using both the verifier's response and a direct comparison of the two agents' atom sets.
 
A document is initially labeled \textbf{GENUINE\_AGREEMENT} when the Implementer's method verifies against the Specifier's predicate on the initial attempt. Despite the label, this indicates only that verification succeeded on the available check. It is treated as provisional agreement pending the mutation and structural audits described below, and we report a corrected rate once those audits are applied. A document is a \textbf{GENERATION\_FAILURE} when verification fails and the verifier reports unresolved identifiers or type errors, representing a plumbing failure rather than a semantic disagreement, and is excluded from the agreement rates we report.

For all other unresolved documents, we compare the two atom sets directly, restricted to variables that both agents extracted. For each shared variable we compare the \emph{set} of (operator, value) pairs, not a single collapsed value, so that a two-sided range with a missing bound is correctly scored as a coverage gap rather than a spurious contradiction. If, for any shared variable, neither atom set's constraints are a subset of the other's, the document is scored as \textbf{NON\_CONVERGENT\_DIRECT\_CONTRADICTION}, indicating the two agents extracted genuinely incompatible thresholds for the same concept. Otherwise it is scored as \textbf{NON\_CONVERGENT\_COVERAGE\_GAP}, meaning the two agents' extracted variables are consistent wherever they overlap, but one agent captured criteria the other missed entirely. Documents where the Specifier's own extraction is empty (\textbf{SPECIFIER\_EMPTY}) or where an agent's output triggers a detected degenerate repetition loop are excluded from the semantically-evaluated denominator, since neither reflects a comparison between two completed formalizations.

\subsection{Mutation Adequacy and Structural Auditing}

Verification success certifies that the Implementer's code satisfies the Specifier's predicate, but it does not certify that the predicate itself is a meaningful discriminator, nor that the Implementer's code was derived independently. We run both checks as a single Stage 3 pipeline on every document reaching GENUINE\_AGREEMENT. First, an anti-cheating structural guard inspects the Implementer's method body for a direct textual invocation of \texttt{MatchSpec}. If present, the document is immediately flagged, since this shortcut trivially discharges the postcondition regardless of whether the Implementer's own independent extraction agrees with the Specifier's at all. Only if this guard passes do we proceed to mutation testing~\cite{black2000mutation,cordy2023mutation}. Synthetic mutants of the specification's atom-derived logic, including tautological, contradictory, boundary-shifted, and operator-inverted variants, are generated and subjected to verification, and the specification is judged adequate only if every mutant is rejected by the verifier. We additionally report structurally audited cases separately as verification-gaming cases and subtract them from the initial agreement count to obtain a corrected agreement rate.

\subsection{Model Configurations}

We evaluate four Specifier/Implementer pairings on the same 96-document high-tier Chia subsample. These include (1) GPT-4.1-mini/GPT-4.1-mini as a same-model control, isolating disagreement that persists even with no capability gap; (2) GPT-4.1-mini/GPT-4.1-nano and (3) GPT-4.1-nano/GPT-4.1-mini, which reverse which role the stronger model occupies; and (4) GPT-5.6-luna/GPT-5.4-mini, the strongest tested configuration. All four pairings use the identical schema, extraction prompts, formalization protocol, and categorization pipeline described above, so that the configurations differ only in the model assigned to each of the two fixed roles.

\section{Experimental Setup}
\label{sec:setup}

We run the protocol described in Section~\ref{sec:method} on the same fixed 96-document high-tier subsample of the Chia corpus across four Specifier/Implementer configurations. These include a same-model control (GPT-4.1-mini/GPT-4.1-mini), the two role-reversed pairings of GPT-4.1-mini and GPT-4.1-nano, and the GPT-5.6-luna/GPT-5.4-mini pairing. All documents, schema, extraction prompts, and the categorization pipeline are held fixed across configurations. Only the model assigned to each of the two fixed roles, meaning the literal Specifier on Route A and the interpretive Implementer on Route B, changes. For each configuration we report the raw category breakdown over all 96 documents, the semantically evaluated rates computed after excluding \texttt{GENERATION\_FAILURE}, \texttt{SPECIFIER\_EMPTY}, and detected degenerate repetition cases from the denominator, the Implementer's mean coverage of the Specifier's extracted variables, document complexity as measured by the Specifier's atom count, the outcome of the mutation-adequacy pass on every \texttt{GENUINE\_AGREEMENT} document, and the outcome of the structural verification-gaming audit, which yields a corrected agreement rate wherever it flags a trivial \texttt{MatchSpec}-invocation case.

\section{Results}
\label{sec:results}

\subsection{Same-model control: agreement is nontrivial even absent a capability gap}

With GPT-4.1-mini serving as both Specifier and Implementer, verification succeeds on the first attempt in 54 of 96 documents. This translates to 56.2\% of all attempted documents, or 59.3\% of the 91 semantically evaluated after excluding 4 generation failures and 1 degenerate-repetition extraction. The remaining semantically evaluated documents split between direct contradiction at 26.4\% and coverage gaps at 14.3\%. Because both roles are filled by the same model, this 59.3\% figure establishes a baseline for independent formalization agreement without a capability asymmetry between Specifier and Implementer. It shows that substantial disagreement persists even when the same model is used in both roles, with 26.4\% of semantically evaluated documents classified as direct non-convergence. All 54 initially agreed specifications passed the full mutation suite, yielding a 100\% adequacy rate, and the structural audit found zero trivial \texttt{MatchSpec}-invocation cases. This baseline therefore reflects substantive agreement rather than an artifact of either verification shortcut or mutation inadequacy.

\begin{table}[h]
\centering
\caption{Cross-agent agreement across four Specifier/Implementer configurations on the 96-document Chia subsample. Rates are computed over the semantically evaluated denominator, excluding \texttt{GENERATION\_FAILURE}, \texttt{SPECIFIER\_EMPTY}, and detected degenerate-repetition cases. Corrected agreement subtracts structurally audited verification-gaming cases.}
\label{tab:main-results}
\begin{tabularx}{\textwidth}{@{}Xccccc@{}}
\toprule
Specifier / Implementer & $N_{\text{eval}}$ & Agreement & Coverage-gap & Contradiction & Corrected \\
\midrule
GPT-4.1-mini / GPT-4.1-mini (control)   & 91 & 59.3\% & 14.3\% & 26.4\% & 59.3\% \\
GPT-4.1-nano / GPT-4.1-mini (weak spec)  & 83 & 53.0\% & 27.7\% & 19.3\% & 53.0\% \\
GPT-4.1-mini / GPT-4.1-nano (weak impl)  & 91 & 14.3\% & 61.5\% & 24.2\% & 13.2\% \\
GPT-5.6-luna / GPT-5.4-mini             & 96 & 34.4\% & 22.9\% & 42.7\% & 34.4\% \\
\bottomrule
\end{tabularx}
\end{table}

\subsection{Agreement is highly sensitive to how capability is allocated across roles}

Holding the model pair (GPT-4.1-mini, GPT-4.1-nano) fixed and swapping which model occupies which role produces a large change in outcome. When the stronger model serves as the Implementer and the weaker model serves as the Specifier, semantically evaluated agreement reaches 53.0\%, which sits close to the same-model baseline. The dominant failure mode here shifts to coverage gaps at 27.7\% rather than contradiction at 19.3\%. Reversing the roles so that the weaker model becomes the Implementer collapses agreement to 14.3\%, where coverage gaps become overwhelmingly dominant at 61.5\%. 

This difference is also reflected in the Implementer's mean coverage of the Specifier's extracted variables, yielding 76.4\% when GPT-4.1-mini implements against GPT-4.1-nano's specification, versus 44.2\% in the reverse pairing. The corresponding counts of documents with zero coverage are 8 of 83 and 15 of 91, respectively. These coverage statistics use the same semantically evaluated documents as the corresponding agreement rates, while the coverage for each document is computed directly from the two extracted atom sets and does not require successful Dafny verification. Coverage is computed directly from the two extracted atom sets and does not require successful Dafny verification. Because the aggregation runs apply the same semantic exclusions used to define each configuration's evaluation set, these coverage statistics are directly interpretable within their respective configurations but should not be treated as interchangeable with the $N_{\text{eval}}$ figures used for agreement rates.

Reversing which role the stronger model occupies therefore produces roughly a $3.7\times$ change in agreement. Because the model pair and all other protocol components are held fixed, this result demonstrates strong sensitivity to the assignment of model capability across the two asymmetric formalization roles. The weak-Implementer configuration also shows the only case in the entire study where verification succeeded through an audited shortcut rather than independent reconstruction, while the weak-Specifier configuration contains one mutation-inadequate agreed specification. These observations underscore that the two roles can fail in qualitatively different ways. Across three independent stochastic runs, the aggregate conclusions were stable: mean agreement was 58.8\% for the same-model control, 53.9\% when GPT-4.1-mini served as Implementer, 12.4\% when GPT-4.1-nano served as Implementer, and 34.3\% for GPT-5.6-luna/GPT-5.4-mini; replicate-level statistics are reported in Appendix~\ref{app:replicates}.

\subsection{The strongest tested configuration remains below the same-model baseline and is contradiction-dominated}

GPT-5.6-luna as Specifier paired with GPT-5.4-mini as Implementer achieves 34.4\% semantically evaluated agreement. No documents were excluded as non-semantic in this configuration. This is higher than the weak-Implementer GPT-4.1 pairing at 14.3\% but remains well below both the strong-Implementer GPT-4.1 pairing at 53.0\% and the same-model GPT-4.1-mini control at 59.3\%. The failure profile is qualitatively distinct from the other three configurations since contradiction at 42.7\% dominates over coverage gaps at 22.9\%. Conversely, every GPT-4.1-family configuration we test has contradiction as, at most, a secondary failure mode relative to either agreement or coverage gaps. Mean Implementer coverage of the Specifier's variables is comparable to the strong-Implementer GPT-4.1 pairing at 83.5\% versus 76.4\%, which is consistent with substantial overlap in the variables extracted by the two agents despite differences in their threshold decisions. All 33 genuinely agreed specifications passed mutation adequacy and none showed evidence of the \texttt{MatchSpec}-invocation shortcut. The 34.4\% figure therefore reflects substantive rather than artifactual agreement. It represents substantive agreement on a minority of documents, while the majority is classified primarily as direct contradiction and, to a lesser extent, coverage gaps.

\subsection{Agreement decreases with specification complexity}

The same-model control also shows a strong relationship between specification complexity and agreement. Using the Specifier's atom count as a document-complexity proxy, agreement decreases from 78.6\% for documents with 1--3 atoms (33 of 42) to 52.8\% for documents with 4--7 atoms (19 of 36), and to 15.4\% for documents with 8 or more atoms (2 of 13). Thus, increasing the number of independently formalized constraints is associated with a substantial reduction in first-attempt agreement. The effect is not limited to a uniform reduction in successful verification because non-convergent outcomes increasingly dominate as specifications become more complex. 

This pattern is also visible in the weak-Implementer configuration, where agreement falls from 26.7\% for 1--3 atoms (12 of 45) to 2.9\% for 4--7 atoms (1 of 34), and to 0\% for 8 or more atoms (0 of 12), where coverage gaps account for 100\% of the 8+ atom cases. These results are consistent with increasing specification complexity making independent reconstruction more difficult, particularly for the weaker Implementer.

\subsection{Verification-gaming audit: one case of agreement through a structural shortcut}
\label{sec:gaming-result}

The structural anti-cheating guard, applied to every \texttt{GENUINE\_AGREEMENT} document across all four configurations, flags exactly one case, specifically \texttt{NCT02186782} in the weak-Implementer GPT-4.1 configuration. In this instance, the Implementer's method body directly invoked \texttt{MatchSpec} rather than reconstructing logic from its own independent extraction. The generated method was:

\begin{verbatim}
method CheckEligibility(age: int, bmi: real) returns (eligible: bool)
  ensures eligible == MatchSpec(age, bmi)
{
    // Based on independent reading no numeric thresholds are specified.
    // Therefore we do not impose any restrictions on age or bmi.
    eligible := MatchSpec(age, bmi);
}
\end{verbatim}

The postcondition therefore verifies trivially, but the method does not implement an independently derived eligibility condition. The accompanying comment is also revealing because rather than deriving logic from the Implementer's own extraction, the generated method delegates the decision directly to the Specifier's predicate, whose thresholds and logic were not provided to the Implementer. The structural audit therefore flags the case as verification-gaming rather than treating the successful verification as genuine agreement. Removing it lowers agreement in that configuration from 14.3\% to a corrected 13.2\%. No other configuration exhibited this pattern. The shortcut was observed only in the GPT-4.1-mini/GPT-4.1-nano weak-Implementer configuration in this experiment. With a single observed case, this should be interpreted as a concrete failure mode rather than evidence that weaker Implementers generally employ such shortcuts.

\subsection{Mutation adequacy and verification validity}

Mutation adequacy is high across the four configurations but is not uniformly perfect. The same-model GPT-4.1-mini control and the GPT-5.6-luna/GPT-5.4-mini configuration achieve 100\% adequacy, with 54/54 and 33/33 initially agreed specifications passing their respective mutation suites. The GPT-4.1-nano/GPT-4.1-mini configuration achieves 97.7\% adequacy (43/44), while the GPT-4.1-mini/GPT-4.1-nano configuration achieves 92.3\% (12/13). The former contains one over-constrained, unsatisfiable agreed specification, \texttt{NCT02429583}, whereas the latter contains the audited \texttt{MatchSpec}-invocation case, \texttt{NCT02186782}. These cases illustrate why successful postcondition verification alone is insufficient since an apparently successful formalization can still fail an independent adequacy or structural audit.

\section{Limitations}
\label{sec:limitations}

\paragraph{Role and extraction policy are confounded.} In our protocol, the Specifier is always assigned the literal extraction prompt (Route A) and the Implementer is always assigned the interpretive extraction prompt (Route B). Consequently, when we reverse roles between GPT-4.1-mini and GPT-4.1-nano, we simultaneously reverse which model is exposed to which extraction policy. The large agreement gap we observe between the two role-reversed configurations, yielding 53.0\% versus 14.3\%, is consistent with a capability effect, an extraction-policy effect, or some combination of the two, and our design cannot separate these. A fully decoupled ablation, in which each model is evaluated under both Route A and Route B while holding the formalization role fixed, would be needed to isolate the contribution of model capability from the contribution of literal versus interpretive reading style.

\paragraph{No repair loop was exercised.} The implementation supports a bounded repair step in which a Symbolic Repair Agent revises the Implementer's method against a Dafny counterexample before a document is marked unresolved. For the runs reported here, $K_{\text{max attempts}} = 1$, so this repair mechanism was never invoked. Every reported outcome reflects a single, unrepaired attempt. Our results therefore characterize zero-shot cross-agent agreement, not the ceiling reachable if the Implementer were allowed to revise its formalization in light of verifier feedback. It is plausible that a repair loop would recover a meaningful fraction of coverage-gap and boundary-strictness disagreements without closing genuinely contradictory readings, but we did not measure this.

\paragraph{No human-annotated gold labels.} By design, our protocol evaluates agreement between two model-derived formalizations rather than agreement with a ground-truth extraction of each document's eligibility criteria. This is a deliberate choice since it lets us apply the protocol at the scale of 96 documents across four model pairings without human annotation, and lets us cleanly separate genuine agreement from verification artifacts. However, it also means we cannot say which agent, if either, extracted the criteria correctly when the two disagree. A contradiction could reflect the Specifier being right and the Implementer wrong, the reverse, or both being wrong in different ways. The qualitative patterns we identify, including boundary strictness, sign inversion on exclusion criteria, and unit or scale mismatches, are diagnostic of where disagreement concentrates, not of which reading is correct.

\paragraph{Single dataset, single verification target.} All results are drawn from one 96-document subsample of the Chia corpus, restricted to criteria that reduce to a single numeric threshold or two-sided range under our schema. Purely categorical or qualitative eligibility criteria, such as required informed consent, are excluded from the extraction task entirely. Our schema's fixed vocabulary means that a criterion that does not map onto any approved variable name cannot be formalized under the current schema, although such out-of-vocabulary cases are explicitly reported by the extraction pipeline. The protocol's behavior on eligibility text that resists this numeric-threshold decomposition, on other domains entirely, or on other formal targets besides Dafny, is untested.

\paragraph{Small, non-exhaustive model set.} We test four Specifier/Implementer pairings across the GPT-4.1 and GPT-5.x model generations. This is sufficient to demonstrate that capability and role allocation interact in a way that a single aggregate accuracy number would hide, but it is not a systematic sweep across model scale, family, or training regime. In particular, we cannot say whether the contradiction-dominated failure profile of the GPT-5.6-luna/GPT-5.4-mini pairing documented in Section~\ref{sec:results} is characteristic of stronger models generally or specific to this pair.

\paragraph{Structural audit targets one known shortcut.} Our verification-gaming audit checks for exactly one failure mode, specifically the literal textual invocation of \texttt{MatchSpec} inside the Implementer's method body. This caught the one instance we observed, but a sufficiently indirect shortcut, for instance an Implementer that reconstructs a formula numerically equivalent to the Specifier's predicate without referencing it by name via some other implicit channel, would not be caught by this guard and would be indistinguishable from genuine agreement in our current pipeline.

\section{Conclusion}
\label{sec:conclusion}

We introduced a cross-agent formalization protocol that treats verification success not as a self-certifying outcome, but as a first-attempt signal that must be checked against structural and mutation-based audits before it is credited as genuine agreement. Applying this protocol to clinical trial eligibility formalization, we showed that a same-model control, which removes any capability gap between the two agents, still produces disagreement on roughly 40.7\% of semantically evaluated documents. This establishes that formalization agreement is not a trivial property of the source text or the schema. We then showed that model capability does not act uniformly across the two roles. Reversing which of two fixed models serves as Specifier versus Implementer changes agreement by more than a factor of three. Furthermore, the strongest model pairing we tested, despite using later-generation models than either GPT-4.1 variant, achieves \emph{lower} agreement than the same-model GPT-4.1-mini baseline and does so through a qualitatively different, contradiction-dominated failure mode. Finally, we demonstrated concretely that verification success alone can mask a structurally dependent shortcut. One case in our weak-Implementer configuration satisfied the Dafny postcondition by invoking the Specifier's predicate directly rather than reconstructing it. This is a pattern our anti-cheating guard was specifically built to catch and that a naive verification-success metric would have silently miscounted as agreement. Together, these results argue that neuro-symbolic pipelines relying on a Specifier/Implementer verification pattern should not treat verifier acceptance as sufficient evidence of independent, meaningful agreement between the two agents, and that both the allocation of capability across roles and the possibility of structural shortcuts deserve explicit auditing rather than being assumed away.

\section{Future Directions}
\label{sec:future}

Several extensions follow directly from the limitations above. Decoupling role from extraction policy by running the literal and interpretive prompts on both members of a model pair in both role assignments would let us separate a genuine capability effect from an artifact of which reading style each model happens to execute more accurately. Enabling and studying the repair loop that our current runs bypass would clarify whether cross-agent disagreement is chiefly a matter of one-shot extraction noise recoverable through verifier-guided revision, or reflects a more persistent disagreement that repair cannot close. A natural follow-up would report agreement as a function of the number of allowed attempts to see whether the same-model baseline saturates quickly while weak-Implementer configurations do not. 

Extending the structural audit beyond literal \texttt{MatchSpec} invocation, for instance by checking whether the Implementer's derived logic is provably equivalent to the Specifier's predicate after broader syntactic normalization rather than only detecting a verbatim reference, would tighten the guarantee that a passing document reflects genuine independent reconstruction. Applying the protocol to other domains with numeric or otherwise formalizable eligibility structure, such as regulatory compliance criteria and financial covenant thresholds, and to other verification-aware targets beyond Dafny would test whether the qualitative failure patterns we observe, including boundary strictness, exclusion-criterion sign inversion, and unit or scale mismatches, are specific to clinical eligibility text or reflect a more general property of independent LLM formalization from natural language. 

Finally, a systematic sweep across a larger and more diverse set of model pairs, varying capability gap size and direction independently, would let us test whether the capability-role interaction we observe generalizes as a predictable function of relative capability, or whether the contradiction-dominated profile of our strongest pairing is idiosyncratic to that particular pair of models.

\section{Broader Impact Statement}
\label{sec:broader-impact}

This work studies clinical trial eligibility criteria as a formalization testbed and does not deploy any system for real clinical trial matching, patient screening, or treatment decision-making. All extracted atoms and generated Dafny artifacts in this paper are used solely to measure agreement between two LLM agents' independent readings of publicly available clinical trial eligibility text from the Chia corpus. None are used to determine any individual's actual trial eligibility. That said, the underlying pattern we study, meaning a Specifier/Implementer pipeline whose verification success is trusted as evidence of correctness, is directly relevant to real deployments of LLM-assisted clinical trial matching tools, where a verified-but-ungrounded formalization could plausibly misclassify a patient's eligibility with real consequences. Our central finding, that verification success is not sufficient evidence of independent agreement and can in fact be satisfied by structurally trivial shortcuts, is intended as a caution against deploying such pipelines on the strength of verifier acceptance alone. This is particularly critical in domains where errors carry direct harm to patients. We see the protocol introduced here, including its mutation-adequacy and structural-auditing components, as a template for the kind of auditing such systems should undergo before any such deployment, rather than as a component to be inserted directly into a clinical pipeline. We do not identify a material dual-use risk specific to this work because the method audits and stress-tests existing model behavior rather than introducing new model capability, and the eligibility-criteria data used are drawn entirely from an existing public research corpus~\cite{kury2020chia}.

\newpage
{\small
\bibliographystyle{unsrt}
\bibliography{references}
}

\newpage

\appendix

\section{Prompts and Reproducibility Details}
\label{app:prompts}

\subsection{Dual-Route Extraction Prompts}
\label{app:extraction-prompts}

Both extraction routes use the same fixed vocabulary and output-format contract but differ in their prescribed reasoning style. Route A is explicitly instructed to perform a strict, literal reading, whereas Route B is instructed to reason about the clinical intent of the eligibility criterion.

The shared format contract requires every emitted value to begin with an explicit comparison operator from \texttt{>=}, \texttt{<=}, \texttt{>}, \texttt{<}, or \texttt{==}. This requirement also applies to implicit-direction cases for which the model must infer the direction needed to express the criterion as a numeric comparison. The prompts restrict extraction to criteria that can be represented by a single numeric threshold or a two-sided numeric range, while purely categorical or qualitative criteria are omitted.

\paragraph{Route A: literal extraction.}
The Route A system prompt begins with the statement that you are a precision clinical data extraction engine performing a strict, literal reading. The prompt provides the complete approved vocabulary and then instructs the model to preserve the boundary exactly as written. In particular, strict and inclusive inequalities must remain distinct, meaning greater than is represented as $>$, while at least is represented as $\geq$. The prompt explicitly prohibits inference of thresholds not stated in the source, except for the minimal operator-direction inference required by the shared format contract. For exclusion criteria, Route A is instructed to extract the eligibility-framed negation of the literal condition.

\paragraph{Route B: interpretive extraction.}
The Route B system prompt begins with the statement that you are a clinical trial coordinator reasoning about what a patient must actually satisfy to be eligible, based on the protocol's intent. Unlike Route A, Route B is explicitly permitted to reason about clinical intent and infer a specific numeric bound when a criterion implies one even when the bound is expressed descriptively. For exclusion criteria, Route B must reason through the exclusion to the eligibility-framed requirement that a patient must satisfy to avoid exclusion. Both routes return only a raw JSON mapping from approved variable names to constraint strings, while markdown formatting and commentary are prohibited.

\subsection{Shared Output Contract}
\label{app:shared-contract}

The two routes share a set of output constraints. First, every emitted value must begin with one of \texttt{>=}, \texttt{<=}, \texttt{>}, \texttt{<}, or \texttt{==}. Second, if a concept matches the meaning of an approved variable name, that exact vocabulary name must be used since models may not substitute an alternative descriptive name. Third, only criteria reducible to a single numeric threshold or a two-sided numeric range are emitted. Fourth, purely categorical or qualitative criteria are omitted. Finally, values must be comparison expressions rather than descriptive English text strings.

The extraction request supplied to both routes is the same user message instructing the model to isolate clinical values from the given text passage. The API request uses structured JSON output through the json\_object type definition. The implementation creates separate client objects for Route A and Route B, using separate endpoint, API-key, API-version, and deployment environment variables.

\subsection{Response Handling and Failure Detection}
\label{app:response-handling}

The implementation performs several deterministic checks before accepting an extraction response. Empty responses are treated as failures. Responses containing markdown code fences are stripped before JSON parsing. A degenerate-repetition detector searches the raw response for repeated key value JSON pairs and marks the response as a repetition failure when the same pair occurs at least five times. Invalid JSON is also reported as an explicit extraction failure rather than silently repaired.

A temperature of $0.2$ was used in the actual API call where ever supported.

\subsection{Compute Resources and Inference Accounting}
\label{app:compute-resources}

All experiments were executed using a local machine equipped with an Intel Core i7 processor running at 1.8~GHz and 16~GB of RAM. No GPU was used for local execution. Local computation was limited to experiment orchestration, document processing, extraction normalization, Dafny verification, mutation testing, structural auditing, and result aggregation. LLM inference was performed through externally hosted model APIs.

Across the four evaluated model configurations, the study generated 3,494 hosted model inference requests and approximately 5.44 million tokens in total. This comprised approximately 5.01 million prompt tokens and 0.43 million completion tokens. The model-specific inference volumes are summarized in Table~\ref{tab:compute-accounting}.

\begin{table}[h]
\centering
\caption{Hosted LLM inference accounting across the four evaluated model configurations. Token counts are approximate and reflect the recorded prompt and completion usage.}
\label{tab:compute-accounting}
\begin{tabular}{lrrrr}
\toprule
Model & Requests & Prompt tokens & Completion tokens & Total tokens \\
\midrule
GPT-4.1-nano & 1,030 & 1.51M & 0.110M & 1.62M \\
GPT-4.1-mini & 2,080 & 2.94M & 0.252M & 3.19M \\
GPT-5.4-mini & 192   & 0.287M & 0.018M & 0.305M \\
GPT-5.6-luna & 192   & 0.272M & 0.048M & 0.320M \\
\midrule
\textbf{Total} & \textbf{3,494} & \textbf{5.01M} & \textbf{0.428M} & \textbf{5.44M} \\
\bottomrule
\end{tabular}
\end{table}

The total hosted inference cost for the reported experiments was approximately 5~USD. Because inference was performed through externally hosted services, the GPU model, GPU-hours, and provider-side hardware utilization were not directly observable or controlled by the authors. Accordingly, we report request and token volumes and the observed inference cost rather than estimating provider-side computational resources.

The local evaluation pipeline did not require GPU acceleration. The reported resource accounting therefore covers the computational resources directly controlled by the experimental setup and the externally observable inference usage of the hosted models.

\subsection{Licenses and Terms for Existing Assets}
\label{app:licenses}

The experiments use the Chia corpus of annotated clinical trial eligibility criteria introduced by Kury et al. The Chia dataset is publicly distributed through Figshare under the Creative Commons Attribution 4.0 International license. The dataset contains free text eligibility criteria and associated annotations from clinical trials registered in ClinicalTrials.gov. The original Chia publication identifies the dataset and provides its Figshare DOI, \texttt{10.6084/m9.figshare.11855817}. The dataset is cited in the main paper and is used here as an existing research corpus rather than as an asset created by this work.

The experiments also rely on externally hosted LLM services and standard software dependencies. These services and dependencies remain subject to their respective provider terms of service and software licenses. No claim of ownership or relicensing of external assets is made by this work.

\section{Canonical Variable Vocabulary and Normalization}
\label{app:normalization}

\subsection{Shared Vocabulary and Type Schema}
\label{app:vocabulary}

The extraction module loads the approved variable names and integer-type metadata from \texttt{final\_unified\_schema.json}. The complete vocabulary is passed to both extraction prompts and is also enforced independently during normalization.

Vocabulary membership is therefore checked twice at different stages because models are instructed to use only approved names, and the deterministic normalizer rejects any returned variable not present in the approved vocabulary. Out-of-vocabulary variables are reported explicitly with their raw value and the reason noting they are not in the shared vocabulary, ensuring they are not silently discarded.

\subsection{Canonical Atom Representation}
\label{app:atoms}

Each accepted constraint is converted into a canonical representation tracking the variable name, operator, value, and integer status. The normalization layer recognizes either a single inequality or a two-sided range. For example, a range string indicating greater than or equal to 40 and less than or equal to 70 is converted into two separate atoms for the same variable. Numeric strings may contain commas and optional units. Integer status is determined from the variable-type map in the shared schema rather than inferred from the emitted value.

Values that do not match the canonical inequality grammar are reported as normalization failures. In particular, disjunctive expressions containing the word or are explicitly reported as unsupported by the normalizer rather than being silently converted into an atom.

\subsection{Degenerate and Malformed Extraction}
\label{app:malformed-extraction}

The extraction pipeline distinguishes several failure modes that occur before semantic comparison. These include empty model responses, invalid JSON, degenerate repetition, out-of-vocabulary variables, null values, and constraint strings that do not conform to the canonical inequality grammar. This separation allows malformed generation and vocabulary violations to be excluded from the semantic agreement denominator without conflating them with genuine disagreements between two valid formalizations.

\section{Evaluation Denominators and Classification Rules}
\label{app:denominators}

\subsection{Semantic Evaluation Denominator}
\label{app:semantic-denominator}

Agreement, coverage-gap, and contradiction rates are computed over the semantically evaluated documents for each model configuration. Documents with extraction or generation failures, empty Specifier output, or detected degenerate repetition are excluded from this denominator. The resulting $N_{\mathrm{eval}}$ therefore differs across configurations.

The same exclusions are applied consistently when reporting the primary semantic outcome rates. This distinction is important because extraction coverage is computed directly from the normalized atom sets and does not require successful Dafny verification.

\subsection{Agreement Classification}
\label{app:agreement-classification-details}

A document is provisionally classified as \texttt{GENUINE\_AGREEMENT} when the independent formalizations satisfy the required Dafny verification condition. Agreement is subsequently subjected to mutation adequacy and structural auditing before being counted as corrected genuine agreement.

For non-agreed documents, the normalized atom sets are compared on shared variables. A direct contradiction is identified when the constraint sets for the shared variables are mutually non-subsuming, meaning neither formalization's constraints constitute a subset of the other's constraints. When one formalization contains constraints that are absent from the other without such a mutual contradiction, the case is classified as a coverage gap.

The resulting evaluation outcomes therefore comprise three primary semantic categories: \texttt{GENUINE\_AGREEMENT}, \texttt{NON\_CONVERGENT\_COVERAGE\_GAP}, \texttt{NON\_CONVERGENT\_DIRECT\_CONTRADICTION}, together with extraction or generation failures that fall outside the semantic denominator.

\section{Mutation Adequacy and Structural Audit Details}
\label{app:audit-details}

Successful verification of the final Dafny program is not treated as sufficient evidence of independent formalization. Agreed specifications are therefore subjected to a mutation-based adequacy check and a structural anti-cheating audit.

The mutation procedure tests whether the verified specification is sensitive to changes in the encoded eligibility conditions. An agreed specification that does not adequately distinguish the generated mutants is marked mutation-inadequate. This check is intended to detect specifications that verify successfully but do not faithfully encode the intended constraints.

The structural audit separately searches for direct invocation of the Specifier's \texttt{MatchSpec} predicate from the Implementer's generated method. Such an invocation can satisfy the Dafny postcondition without demonstrating independent reconstruction of the eligibility logic and is therefore treated as verification gaming.

The primary agreement result is corrected by removing cases flagged by the structural verification-gaming audit. Mutation inadequacy is reported separately as an independent specification-adequacy diagnostic. The detailed mutation outcomes for each model configuration are reported in the main Results section, while document-level audit traces are retained in the supplementary results.

\section{Extraction-level diagnostics and substantive divergences}

The atom-level audit exposes both extraction-discipline failures and limitations of exact-name alignment that are not visible in the aggregate agreement rates. First, in \texttt{NCT02429583}, the GPT-4.1-nano Implementer violated the shared vocabulary constraint by emitting approximately 18 variables such as \texttt{willing\_to\_receive}, \texttt{volunteer\_infected\_with\_HCV}, and \texttt{health\_status}, none of which appears in the approved vocabulary. Several are not naturally numeric eligibility variables. Because the canonical parser routes out-of-vocabulary variables to an explicit diagnostic report rather than silently accepting them, this constitutes a measurable extraction-discipline failure rather than a downstream script artifact. We therefore treat vocabulary compliance as a distinct extraction-discipline signal rather than folding the failure into semantic disagreement.

Second, exact string matching can make the shared-variable set shallower than the underlying semantic overlap. In \texttt{NCT03335904}, for example, Route A extracts \texttt{apnea\_hypopnea\_index}, whereas Route B extracts \texttt{ahi\_range\_upper}. These names refer to the same underlying concept but are not identical strings. Consequently, the exact-name comparison treats only the shared \texttt{bmi} variable as comparable in that document. The resulting \texttt{GENUINE\_AGREEMENT} classification is correct under the protocol's exact-name alignment rule, but it does not establish that all independently extracted criteria agree. We do not introduce a fuzzy or LLM-based variable aligner because an additional unverified semantic artifact would itself become part of the core measurement procedure. Instead, we report \texttt{shared\_var\_count} alongside the outcome and identify exact-string variable alignment as a limitation. A match over one shared variable therefore has weaker evidentiary coverage than a match over all variables independently extracted by both routes.

The protocol nevertheless identifies substantive divergences without requiring semantic variable alignment. In \texttt{NCT02612181}, the two routes independently produce \texttt{age >= 18} and \texttt{age < 18}, respectively, representing a complete logical reversal of the age criterion. In \texttt{NCT03199560}, they produce \texttt{age > 18} and \texttt{age >= 18}, exposing a strict-versus-inclusive boundary disagreement. These cases illustrate the value of independent dual-route extraction, where disagreements can be detected directly from the formal atoms without requiring an external gold specification, while the nature of the disagreement remains mechanically inspectable.

\paragraph{Omission versus fabrication.}
Coverage gaps also encompass qualitatively different behaviors that the aggregate category does not distinguish. In one GPT-4.1-mini/GPT-4.1-mini case, for example, the Implementer explicitly leaves an unsupported parameter unconstrained:
\begin{verbatim}
var egfr_ok := true; // no information about eGFR constraints
\end{verbatim}
while elsewhere in the same method it introduces an unsupported relationship:
\begin{verbatim}
var life_expectancy_ok := age <= max_life_expectancy_years; // fabricated relation
\end{verbatim}
The first behavior is an explicit omission in the absence of source information, whereas the second introduces a semantic constraint not justified by the extracted criteria. Both can manifest as coverage-related disagreement in the current taxonomy, but they represent different failure modes, specifically conservative incompleteness versus unsupported fabrication. We therefore retain the aggregate coverage-gap category for quantitative analysis while noting this distinction in the qualitative audit.

\section{Representative Formalization Examples}
\label{app:formalization-examples}

\subsection{Contradictory Thresholds}

The protocol permits direct contradictions to be identified after normalization even when the corresponding source language is expressed in different textual forms. For example, the pair

\begin{quote}
\texttt{age >= 18}
\end{quote}
and
\begin{quote}
\texttt{age < 18}
\end{quote}

defines disjoint eligibility conditions and is classified as a direct contradiction.

\subsection{Boundary Sensitivity}

The distinction between

\begin{quote}
\texttt{age > 18}
\end{quote}
and
\begin{quote}
\texttt{age >= 18}
\end{quote}

is preserved throughout normalization. The protocol does not collapse strict and inclusive inequalities because they induce different eligibility sets.

\subsection{Verification Shortcut}

The audited gaming case, \texttt{NCT02186782}, demonstrates why the structural audit is necessary. The Implementer generated a method that directly invoked the Specifier's \texttt{MatchSpec} predicate rather than reconstructing the eligibility logic from its own independent extraction. Although Dafny verification succeeded, the generated implementation did not constitute independent formalization and was therefore removed from corrected agreement.

\subsection{Complexity Measure}

For the complexity analysis in the main Results, document complexity is measured by the number of normalized atoms extracted by the Specifier. A single-threshold criterion contributes one atom, while a two-sided range contributes two atoms. Documents are grouped into atom-count buckets of 1--3, 4--7, and 8 or more atoms. This measure captures the number of independently represented numeric constraints rather than document length or token count.

\section{Replicate Robustness Analysis}
\label{app:replicates}

To assess sensitivity to stochastic variation in model generation, we repeated the complete evaluation protocol three times for each of the four Specifier and Implementer configurations, using the same fixed set of 96 clinical trial documents in every run. Each repetition independently re-executed the model calls and the downstream extraction, formalization, verification, mutation, and structural-audit pipeline. The primary Results section reports the first run for consistency with the document-level analyses, while this appendix reports aggregate statistics across all three runs.

Table~\ref{tab:replicate-robustness} summarizes the resulting agreement, corrected agreement, Implementer coverage, and contradiction rates. The ordering of configurations is consistent across all three runs. The same-model GPT-4.1-mini control achieves $58.8 \pm 2.0\%$ mean agreement, while the role-reversed GPT-4.1 configurations produce substantially different outcomes, yielding $53.9 \pm 3.1\%$ when GPT-4.1-mini serves as Implementer and $12.4 \pm 2.3\%$ when GPT-4.1-nano serves as Implementer. The GPT-5.6-luna/GPT-5.4-mini configuration achieves $34.3 \pm 0.5\%$ agreement. Thus, the central role-allocation result persists across independent stochastic runs rather than depending on a single generation.

The failure profiles are similarly consistent. The weak-Implementer configuration has a mean coverage-gap rate of $63.6 \pm 3.3\%$ and mean Implementer coverage of $43.2 \pm 2.2\%$, compared with $21.4 \pm 5.4\%$ coverage gaps and $81.7 \pm 4.7\%$ coverage for the corresponding strong-Implementer configuration. The GPT-5.6-luna/GPT-5.4-mini configuration remains dominated by direct contradiction, with a mean contradiction rate of $42.3 \pm 1.1\%$. These patterns indicate that the observed differences are not limited to variation in aggregate agreement rates but persist in the qualitative distribution of failure modes. Because all three runs use the same 96 documents, these replicate statistics characterize stochastic variation in model generation rather than sampling uncertainty over documents. We therefore do not perform or report formal hypothesis tests across configurations. The primary conclusions are instead supported by the consistency of the aggregate ordering and failure profiles across repeated runs.

\begin{table}[t]
\centering
\caption{Replicate robustness across three independent stochastic runs on the same 96-document evaluation set. Values are mean $\pm$ standard deviation across runs.}
\label{tab:replicate-robustness}
\begin{tabular}{lcccc}
\toprule
Specifier / Implementer & Agreement & Corrected & Coverage & Contradiction \\
\midrule
GPT-4.1-mini / GPT-4.1-mini & $58.8 \pm 2.0\%$ & $58.8 \pm 2.0\%$ & $88.1 \pm 1.3\%$ & $24.1 \pm 2.9\%$ \\
GPT-4.1-nano / GPT-4.1-mini & $53.9 \pm 3.1\%$ & $53.9 \pm 3.1\%$ & $81.7 \pm 4.7\%$ & $24.7 \pm 5.6\%$ \\
GPT-4.1-mini / GPT-4.1-nano & $12.4 \pm 2.3\%$ & $10.9 \pm 2.3\%$ & $43.2 \pm 2.2\%$ & $24.0 \pm 1.1\%$ \\
GPT-5.6-luna / GPT-5.4-mini & $34.3 \pm 0.5\%$ & $34.3 \pm 0.5\%$ & $84.9 \pm 2.5\%$ & $42.3 \pm 1.1\%$ \\
\bottomrule
\end{tabular}
\end{table}

The standard deviations characterize run-to-run stochastic variation on the same document set and should not be interpreted as uncertainty from independently sampled datasets. We therefore use these replicates as a robustness analysis rather than as a basis for formal significance testing. At the individual-document level, outcomes are less stable than the aggregate rates. The fraction of documents retaining the same categorical outcome across all three runs is 71.9\% for the same-model control, 60.4\% for GPT-4.1-nano/GPT-4.1-mini, 64.6\% for GPT-4.1-mini/GPT-4.1-nano, and 51.0\% for GPT-5.6-luna/GPT-5.4-mini. Thus, stochastic variation can change the categorical outcome of individual documents even when aggregate agreement rates remain comparatively stable.


\newpage
\section*{NeurIPS Paper Checklist}

The checklist is designed to encourage best practices for responsible machine learning research, addressing issues of reproducibility, transparency, research ethics, and societal impact. Do not remove the checklist: {\bf The papers not including the checklist will be desk rejected.} The checklist should follow the references and follow the (optional) supplemental material.  The checklist does NOT count towards the page
limit. 

Please read the checklist guidelines carefully for information on how to answer these questions. For each question in the checklist:
\begin{itemize}
    \item You should answer \answerYes{}, \answerNo{}, or \answerNA{}.
    \item \answerNA{} means either that the question is Not Applicable for that particular paper or the relevant information is Not Available.
    \item Please provide a short (1--2 sentence) justification right after your answer (even for \answerNA). 
\end{itemize}

{\bf The checklist answers are an integral part of your paper submission.} They are visible to the reviewers, area chairs, senior area chairs, and ethics reviewers. You will also be asked to include it (after eventual revisions) with the final version of your paper, and its final version will be published with the paper.

The reviewers of your paper will be asked to use the checklist as one of the factors in their evaluation. While \answerYes{} is generally preferable to \answerNo{}, it is perfectly acceptable to answer \answerNo{} provided a proper justification is given (e.g., error bars are not reported because it would be too computationally expensive'' or ``we were unable to find the license for the dataset we used''). In general, answering \answerNo{} or \answerNA{} is not grounds for rejection. While the questions are phrased in a binary way, we acknowledge that the true answer is often more nuanced, so please just use your best judgment and write a justification to elaborate. All supporting evidence can appear either in the main paper or the supplemental material, provided in appendix. If you answer \answerYes{} to a question, in the justification please point to the section(s) where related material for the question can be found.

IMPORTANT, please:
\begin{itemize}
    \item {\bf Delete this instruction block, but keep the section heading ``NeurIPS Paper Checklist"},
    \item  {\bf Keep the checklist subsection headings, questions/answers and guidelines below.}
    \item {\bf Do not modify the questions and only use the provided macros for your answers}.
\end{itemize}


\begin{enumerate}

\item {\bf Claims}
    \item[] Question: Do the main claims made in the abstract and introduction accurately reflect the paper's contributions and scope?
    \item[] Answer: \answerYes{} 
    \item[] Justification: The abstract and introduction characterize the work as an empirical study of cross-agent agreement in LLM-to-Dafny formalization, including the effects of role assignment, specification complexity, and verification shortcuts. The claims are scoped to the evaluated clinical-trial corpus, model configurations, and independent dual-route protocol rather than being presented as universal properties of LLM formalization.

\item {\bf Limitations}
    \item[] Question: Does the paper discuss the limitations of the work performed by the authors?
    \item[] Answer: \answerYes{}
    \item[] Justification: The paper discusses limitations including the fixed vocabulary and exact-name alignment protocol, the absence of an external gold specification, the limited model configurations and corpus, and the distinction between coverage gaps, omissions, and unsupported fabrication. The Appendix also provides document-level extraction diagnostics illustrating these limitations.

\item {\bf Theory assumptions and proofs}
    \item[] Question: For each theoretical result, does the paper provide the full set of assumptions and a complete (and correct) proof?
    \item[] Answer: \answerNA{}
    \item[] Justification: The paper does not present theoretical results requiring formal assumptions, theorems, lemmas, or proofs. Its contribution is primarily empirical and methodological.

    \item {\bf Experimental result reproducibility}
    \item[] Question: Does the paper fully disclose all the information needed to reproduce the main experimental results of the paper to the extent that it affects the main claims and/or conclusions of the paper (regardless of whether the code and data are provided or not)?
    \item[] Answer: \answerYes{}
     \item[] Justification: The paper specifies the dual-route protocol, model configurations, extraction prompts and output constraints, normalization procedure, evaluation denominators, agreement classification, mutation-adequacy checks, and structural audit. The Appendix provides additional implementation and reproducibility details needed to reconstruct the reported evaluation.

\item {\bf Open access to data and code}
    \item[] Question: Does the paper provide open access to the data and code, with sufficient instructions to faithfully reproduce the main experimental results, as described in supplemental material?
    \item[] Answer: \answerYes{}
    \item[] Justification: The source code and experimental artifacts required to reproduce the reported evaluation are publicly available through an anonymous repository at \url{https://github.com/priyanathmaji/llm-dafny-agreement/}. The Chia clinical-trial eligibility corpus used in the experiments is not redistributed with the submission; instead, the Appendix identifies the original dataset source, licensing information, and instructions for obtaining and placing the dataset locally. No restricted or proprietary dataset is included in the repository.

\item {\bf Experimental setting/details}
    \item[] Question: Does the paper specify all the training and test details (e.g., data splits, hyperparameters, how they were chosen, type of optimizer) necessary to understand the results?
    \item[] Answer: \answerYes{}
    \item[] Justification: The study does not train a new model; it evaluates fixed LLM configurations on a defined 96-document clinical-trial corpus. The paper and Appendix specify the model pairings, dual-route extraction procedure, prompts, vocabulary constraints, normalization rules, semantic evaluation denominator, and verification and mutation procedures.

\item {\bf Experiment statistical significance}
    \item[] Question: Does the paper report error bars suitably and correctly defined or other appropriate information about the statistical significance of the experiments?
    \item[] Answer: \answerYes{}
    Justification: The paper reports run-to-run variability using mean ± standard deviation across three independent stochastic runs for each model configuration. These error bars quantify stochastic variation from repeated model generation on the same fixed 96-document evaluation set. Because the replicates reuse the same documents, they are not treated as independent samples for formal statistical significance testing; accordingly, the paper does not report p-values or claim statistical significance from these replicates. Document-level outcomes and the corresponding aggregate agreement rates are also reported explicitly.

\item {\bf Experiments compute resources}
    \item[] Question: For each experiment, does the paper provide sufficient information on the computer resources (type of compute workers, memory, time of execution) needed to reproduce the experiments?
    \item[] Answer: \answerYes{}
    \item[] Justification: The experiments use externally hosted LLM inference and local execution for orchestration, extraction normalization, Dafny verification, mutation testing, and analysis. The local evaluation environment used an Intel Core i7 processor at 1.8 GHz with 16 GB RAM and no GPU. Across the reported experiments, the evaluation generated approximately 3,494 hosted-model inference requests and 5.44 million tokens in total, comprising approximately 5.01 million prompt tokens and 0.43 million completion tokens. The model-specific totals were 1.62M tokens for GPT-4.1-nano, 3.19M for GPT-4.1-mini, 0.305M for GPT-5.4-mini, and 0.320M for GPT-5.6-luna. The total hosted inference cost was approximately \$5. Local execution did not require GPU acceleration.
    
\item {\bf Code of ethics}
    \item[] Question: Does the research conducted in the paper conform, in every respect, with the NeurIPS Code of Ethics \url{https://neurips.cc/public/EthicsGuidelines}?
    \item[] Answer: \answerYes{}
    \item[] Justification: The work evaluates language-model formalization behavior on a clinical-trial document corpus and does not conduct experiments involving human participants, deploy an automated clinical decision system, or intentionally create harmful capabilities. The study is conducted as an evaluation of formalization reliability and includes explicit auditing for verification shortcuts.

\item {\bf Broader impacts}
    \item[] Question: Does the paper discuss both potential positive societal impacts and negative societal impacts of the work performed?
    \item[] Answer: \answerYes{}
    \item[] Justification: The paper discusses the potential benefit of more reliable formalization of eligibility criteria for applications involving high-stakes or compliance-sensitive information, while also acknowledging risks from incorrect extraction, unsupported fabrication, and false agreement. The study's verification and auditing procedures are presented as safeguards against treating successful formal verification as sufficient evidence of semantic correctness.

\item {\bf Safeguards}
    \item[] Question: Does the paper describe safeguards that have been put in place for responsible release of data or models that have a high risk for misuse (e.g., pre-trained language models, image generators, or scraped datasets)?
    \item[] Answer: \answerNA{}
    \item[] Justification: The paper does not release a new pretrained model, generative model, or other high-risk model artifact. The evaluated LLMs are accessed as externally hosted services rather than released by the authors.

\item {\bf Licenses for existing assets}
    \item[] Question: Are the creators or original owners of assets (e.g., code, data, models), used in the paper, properly credited and are the license and terms of use explicitly mentioned and properly respected?
    \item[] Answer: \answerYes{}
    \item[] Justification: The Chia clinical-trial eligibility corpus used in the experiments is distributed under the Creative Commons Attribution 4.0 International (CC BY 4.0) license. The paper identifies the dataset and its Figshare record, and the dataset metadata specifies the applicable license. The submission cites the original Chia publication and identifies the dataset source in the Appendix.

\item {\bf New assets}
    \item[] Question: Are new assets introduced in the paper well documented and is the documentation provided alongside the assets?
    \item[] Answer: \answerNo{}
    \item[] Justification: The paper introduces an evaluation protocol, prompts, schemas, and analysis artifacts, but these are not currently being released as a separately documented public asset package. The Appendix documents the protocol and implementation details used in the study.

\item {\bf Crowdsourcing and research with human subjects}
    \item[] Question: For crowdsourcing experiments and research with human subjects, does the paper include the full text of instructions given to participants and screenshots, if applicable, as well as details about compensation (if any)? 
    \item[] Answer: \answerNA{}
    \item[] Justification: The study does not involve crowdsourcing experiments or human participants. All evaluation is performed computationally using existing clinical-trial documents and LLM inference.

\item {\bf Institutional review board (IRB) approvals or equivalent for research with human subjects}
    \item[] Question: Does the paper describe potential risks incurred by study participants, whether such risks were disclosed to the subjects, and whether Institutional Review Board (IRB) approvals (or an equivalent approval/review based on the requirements of your country or institution) were obtained?
    \item[] Answer: \answerNA{}
    \item[] Justification: The study does not involve human subjects or collect data directly from participants and therefore does not require participant consent or IRB review for the reported experiments.

\item {\bf Declaration of LLM usage}
    \item[] Question: Does the paper describe the usage of LLMs if it is an important, original, or non-standard component of the core methods in this research? Note that if the LLM is used only for writing, editing, or formatting purposes and does \emph{not} impact the core methodology, scientific rigor, or originality of the research, declaration is not required.
    \item[] Answer: \answerYes{}
    \item[] Justification: LLMs are the central experimental components of the proposed dual-route formalization protocol. The paper identifies the evaluated model configurations and documents their roles, extraction prompts, output constraints, and use in generating Dafny formalizations.

\end{enumerate}

\end{document}